\documentclass[10pt,twocolumn,letterpaper]{article}

\usepackage{cvpr}
\usepackage{times}
\usepackage{epsfig}
\usepackage{graphicx}
\usepackage{amsmath}
\usepackage{amssymb}
\usepackage{graphicx}
\usepackage{subfig}
\newcommand{\tab}[1]{Table~\ref{tab:#1}}
\newcommand{\secc}[1]{Section~\ref{secc:#1}}
\newcommand{\fig}[1]{Figure~\ref{fig:#1}}
\newcommand{\z}{\mathbf{z}}
\newcommand{\dvec}{\mathbf{d}}
\newcommand{\x}{\mathbf{x}}

\newcommand\blfootnote[1]{%
  \begingroup
  \renewcommand\thefootnote{}\footnote{#1}%
  \addtocounter{footnote}{-1}%
  \endgroup
}

\usepackage[pagebackref=true,breaklinks=true,letterpaper=true,colorlinks,bookmarks=false]{hyperref}
\cvprfinalcopy

\def\cvprPaperID{****} 

\ifcvprfinal\fi
\begin{document}

\title{Image Counterfactual Sensitivity Analysis for DetectingUnintended Bias}

\author{Remi Denton \\Ben Hutchinson \\Margaret Mitchell \\Timnit Gebru \\Andrew Zaldivar\\
{\tt\small \{dentone, benhutch, mmitchellai, tgebru, andrewzaldivar\}@google.com }
}
\maketitle

\begin{abstract}

    Facial analysis models are increasingly used in applications that have serious impacts on people's lives, ranging from authentication to surveillance tracking. It is therefore critical to develop techniques that can reveal unintended biases in facial classifiers to help guide the ethical use of facial analysis technology. This work proposes a framework called \textit{image counterfactual sensitivity analysis}, which we explore as a proof-of-concept in analyzing a smiling attribute classifier trained on faces of celebrities.  The framework utilizes counterfactuals to examine how a classifier's prediction changes if a face characteristic slightly changes. We leverage recent advances in generative adversarial networks to build a realistic generative model of face images that affords controlled manipulation of specific image characteristics. We then introduce a set of metrics that measure the effect of manipulating a specific property on the output of the trained classifier. Empirically, we find several different factors of variation that affect the predictions of the smiling classifier. This proof-of-concept demonstrates potential ways generative models can be leveraged for fine-grained analysis of bias and fairness.
\end{abstract}

\blfootnote{CVPR 2019 Workshop on Fairness Accountability Transparency and Ethics in Computer Vision.}

\section{Introduction}

Facial analysis models are increasingly being used to make
high stakes decisions, including border patrol~\cite{rudolph2017}, police surveillance, tracking of
protesters~\cite{harvie2019}, and assisting hiring decisions~\cite{raghavan2019}, as well as in consumer
applications such as automated photo tagging and
automated phone unlock.
Studies have demonstrated that in the US, the negative consequences of
deploying systems in high stakes scenarios,
which includes infringing on individual dignity, autonomy, and privacy,
will be borne disproportionately by  communities of color~\cite{harvie2019}.
In light of this, there is a growing
need for evaluations of model bias to complement processes that address whether and how to deploy ML models.

A range of previous studies have uncovered systematic
demographic biases in the performance of computer vision systems. For example, in 2012, \cite{klare2012} found that
facial recognition models performed worse on female
faces, black faces, and younger adults.
More recently, an analysis of commercial gender classification systems demonstrated disparities with respect to perceived gender and skin type, with darker-skinned females exhibiting the highest error rates  \cite{gendershades, Raji2019}.
An analysis of state-of-the-art object detection systems found decreased pedestrian detection accuracy on darker skin tones
\cite{Wilson2019}. Gender-based bias has also been observed in image captioning \cite{10.1007/978-3-030-01219-9_47} and image classification \cite{Zhao2017}.

We propose a novel framework for identifying biases of a face attribute classifier. Specifically, we consider a classifier trained to predict the presence or absence of a smile in a face image. However, the approach generalizes to a broad range of image classification problems. Our method, which we refer to as \textit{image counterfactual sensitivity analysis}, tests how the prediction of a trained classifier changes if a characteristic of the image is different. In this paper, we apply the method to a smiling classifier, posing questions of the form: "Would the smiling prediction change if this person's hair had been longer?"

To perform these analyses, we
introduce a method that we call \textit{counterfactual face attribute manipulation}.
Our method leverages recent advances in generative models and post-hoc interpretability to facilitate an ethics-informed audit exploring normative constraints on what \textit{should} be independent of the smiling attribute.
First, we build a directed generative model of faces that maps latent codes, sampled from a fixed prior distribution, to images.
Next, we infer directions in latent code space that correspond to manipulations of a particular image attribute in pixel space. We refer to these vectors as \textit{attribute vectors}. Traversing these directions in latent space provides a straightforward method of generating images that smoothly vary one characteristic of a face, while preserving others. We propose several metrics that measure the sensitivity of the trained classifier to manipulations imposed by attribute vectors.

This paper has several novel contributions. First, we introduce the technique of \textit{counterfactual face attribute manipulation}, which uses a Generative Adversarial Network and an image encoder in order to learn and manipulate latent representations of face attributes.
Second, we adapt techniques from the field of
sensitivity analysis---related to stability analysis \cite{bousquet2002}---to the task of auditing
trained ML models. We define two
model auditing techniques, which we call
\textit{score sensitivity} and
\textit{classification sensitivity}.
Third, the combination of counterfactual face manipulation and sensitivity analyses
comprises a novel method for the
counterfactual testing
of facial analysis models for undesirable bias.
We test our method on a classifier trained to predict the smiling attribute from faces in the CelebA dataset \cite{celeba} and find our approach reveals several biases in the smiling classifier.
Last, we discuss the challenges and limitations of testing on synthetically generated data sets, and outline a research
program that systematically includes humans in the loop in order to evaluate the societal validity
of processes for generating counterfactual data.


\section{Ethical Considerations}\label{secc:ethical}
\textbf{Intended use and limitations:} The techniques proposed in this paper can be applied to detect and analyze unintended and undesirable biases in a wide variety of face-centric computer vision systems. While this type of analysis is an important part of designing fair and inclusive technology, it is not sufficient. Rather, it must be part of a larger, socially contextualized project to critically assess ethical concerns relating to facial analysis technology.
This project must include addressing questions of whether and when to deploy technologies,
frameworks for democratic control and accountability, and design practices which emphasize autonomy and privacy.

In the experiments described below in Section~\ref{Results}, we focus our attention on smiling detection because we believe it has a range of beneficial applications and limited harmful applications, including aiding in the selection of images from a stream of images,
ML-assisted photography, and
augmented reality applications where a smile triggers added features.
And we take care to distinguish the facial expression of smiling from the emotion of happiness, since for this
inference of emotion from expression to be justifiable would require inferential reliability, specificity,
generalizability, and validity \cite{barrett2019}.
We note that although cultures differ in the social-functional associations of smiling \cite{rychlowska2015},
the facial expression itself does seem to be a cross-cultural universal, and hence
smiling detection does not risk exposing any personal information about social
groups or personal identities.
On the other hand, there are many facial analysis systems that do have harmful applications where foreseeable risk outweighs foreseeable benefit, e.g., where the facial analysis task itself has the potential to perpetuate, reinforce and amplify societal injustices, regardless of how balanced the classification performance is with respect to different demographics. For example, face recognition and gender classification technology represent two facial analysis domains with a great potential for abuse, with the highest risks often falling on already marginalized populations \cite{Keyes2018, annon, msrfacerec}.


\textbf{CelebA dataset:}
We use the CelebA dataset \cite{celeba} and its attribute annotations in the experiments described
below in Section~\ref{Results}. As this dataset contains public domain images of public figures, it avoids the issues of some other public domain datasets of face images, e.g., \cite{klare2012}.
All of the attributes within the dataset are operationalized as binary categories (i.e., the attribute is present or it is absent). We note that in many cases this binary categorization does not reflect the real human diversity of attributes.
This is perhaps most clear when the attributes are related to continuous factors such as color of hair, skin or lips, waviness of hair, age, and size of nose or lips. However it also applies to almost all the attributes with the exception of \texttt{Wearing\_Earrings}, \texttt{Wearing\_Hat}, \texttt{Wearing\_Necklace} and \texttt{Wearing\_Necktie}.
We also note that the attributes \texttt{Attractive} and \texttt{Chubby} involve normative judgments that
may cause harm when interpreted perjoratively.
Both the continuous and subjective nature of the attributes make category boundaries contingent on the annotators and annotation instructions.
As a result, the factors of variation that we manipulate in our experiments are tied to the ways the attributes have been operationalized and annotated within the CelebA dataset.

After careful consideration, we have chosen not to manipulate images based on the \texttt{Male} attribute. While this attribute may, on the surface, appear relevant to fairness evaluations, the risk and potential for negative impact outweighs the perceived benefit of examining this attribute in  particular.
First, the CelebA annotation scheme conceptualizes the \texttt{Male} attribute as both binary and perceptually obvious and discernible. Using a gender binary in reporting results implicitly condones the classification of gender into two distinct and opposite categories, which perpetuates harm against individuals who exist outside the bounds of this categorization scheme and reinforces rigid social norms of gender expression.
Second, due to celebrity composition of this dataset, a very narrow range of gender expression is reflected in the images. Consequently, applying our methodology to the \texttt{Male} attribute would result in image manipulations that alter a specific set of correlated image features that---through a largely cisnormative and heteronormative cultural lens---are commonly interpreted as gender signifiers. Even with a nuanced discussion of what the \texttt{Male} attribute vector is (and is not) encoding, presenting these results risks reinforcing and implicitly condoning the views that individuals of different genders \textit{do} and \textit{should} present a certain way.
Generally speaking, we recommend researchers avoid attempts to manipulate images based on unstable social constructs like gender. Instead, we suggest limiting manipulations to attributes defined by observable facial characteristics. 

\begin{figure}[t!]
  \centering
    \includegraphics[width=\linewidth]{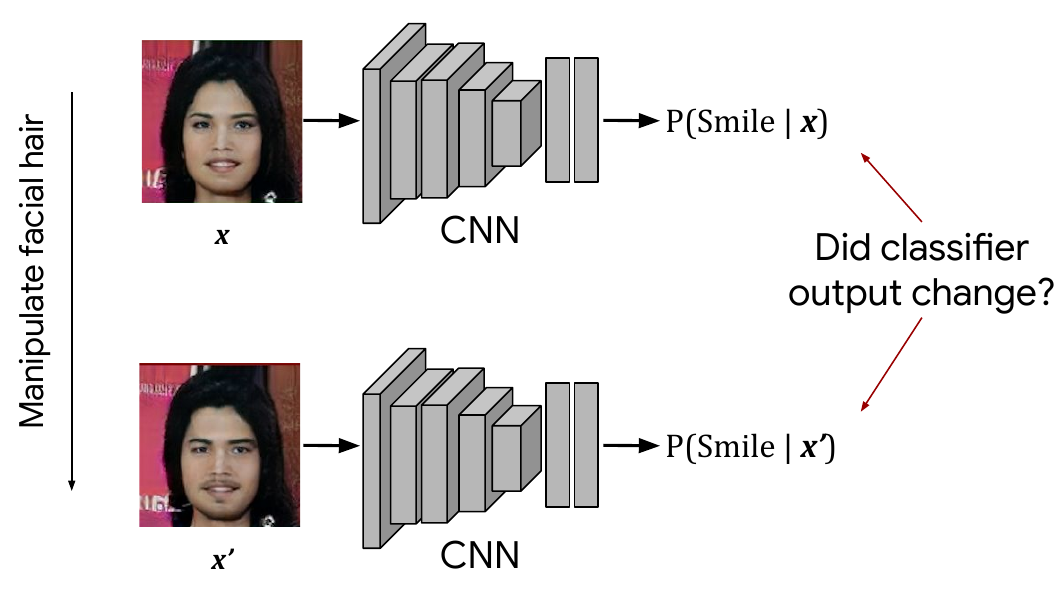}
  \caption{Counterfactual attribute sensitivity measures the effect of manipulating a specific property of an image on the output of a trained classifier. In this example, we consider the effect of adding facial hair on the output of a smiling classifier. If the classifier's output systematically changes as a result of adding or removing facial hair then a potentially undesirable bias has been detected since, all else being equal, facial hair should be irrelevant to the classification task. }
\end{figure}
\label{fig:example}

\section{Related work}

Several works have focused on evaluating the fairness of human-centric vision classifiers. \cite{gendershades, Raji2019} compare error statistics of facial analysis systems across groups defined by Fitzpatrick skin type \cite{Fitzpatrick} and gender;
while \cite{klare2012} compares across groups defined by race/ethnicity, age and gender, as
categorized by Florida police records.
\cite{Mitchell2019ModelCF} proposes standardized model reporting guidelines that include evaluations  disaggregated by groups defined along cultural, demographic, and phenotypic lines. Our work is complementary to disaggregated evaluations based on real images, while incorporating recent work in post-hoc interpretability for understanding biases. Specifically, the generative techniques we propose allow new types of evaluation not easily afforded by real image datasets: (i) We can assess how a prediction might change if characteristics of an image \textit{that we believe should be irrelevant to classification} are altered in a specific targeted manner. This means we can identify a causal relationship between features in an image and the classifier's output; (ii) Our approach facilitates testing on faces with combinations of characteristics that may be underrepresented in the testing set.

There has been significant research effort devoted to developing post-hoc approaches for interpreting and explaining neural networks (e.g. see \cite{Gilpin2018} for a review).
The most common framework involves computing saliency maps to visualize the importance of individual pixels to a classifier's prediction. Gradient-based saliency approaches \cite{Simonyan2013DeepIC, Sundararajan2017, Selvaraju} attribute importance based on first-order derivatives of the input. Another set of approaches determine importance based on perturbing individual or groups of pixels and observing the change in classifier output \cite{Chang2018ExplainingIC,Fong2017InterpretableEO, Dabkowski2017RealTI}. The perturbation methods relate to our approach in that counterfactual-style inputs drive the search for regions that most affect the classifier's output. Our method is different in that, instead of searching over perturbations that maximally influence the classifiers output, we perturb the input along known and meaningful factors of variation to test the classifiers response.

Testing with Concept Activation Vectors (TCAV) \cite{tcav} is an alternative interpretability framework that poses explanations at the level of high level concepts rather than pixels. TCAV defines a concept via a set of data instances (e.g., images) sharing a common characteristic. Similarly,  our method infers attribute  vectors based on two sets of images: (i) images with the attribute and (ii) images without the attribute. TCAV uses directional derivatives to reveal the `conceptual sensitivity' of a model's prediction of a class (e.g \texttt{Smiling}) to a concept by the two sets of images). Our approach extends this technique to measure the effect of an attribute vector \textit{directly on the model predictions}. TCAV provides a general framework for model interpretability, and our approach narrows in on cases where generative techniques produce high quality and perceptually realistic images. In the settings where our method is applicable, it provides  direct and actionable information about a model's decisions. We believe the two approaches are complementary  and can work in combination: TCAV can point to a potential bias, and our method can provide clear interpretable evidence of bias in classifier predictions.

Our sensitivity measures are related to the fields of perturbation analysis and sensitivity analysis
(e.g., \cite{cacuci2005});
in Machine Learning these are known as stability analysis when applied to the model training process
\cite{bousquet2002}.

Our method is related to  \cite{Kilbertus2017, Kusner2017}, who propose the notion of counterfactual fairness in the context of causal modeling. Here, the fairness of a model is determined by assessing the effect of  counterfactual interventions on a sensitive attribute (e.g., race or gender) over a causal graph. Recently, \cite{Kohler-Hausmann, moss2019} have argued this framework is based on a misunderstanding of how race operates in our society and the flawed assumption that social groups like race and gender are entities that can have counterfactual causality. Crucially, our approach manipulates observable characteristics of a face, rather than conceptualizing the manipulation at the level of social groups. Of course, some of these attributes (e.g.,~presence of a beard) may be strongly correlated with a particular social group (e.g.~gender identities), but they neither constitute nor are determined by the group.

\cite{Sahaj2019} considers counterfactual fairness in text classification, proposing training and evaluation techniques based on counterfactual manipulation of terms referencing different identify groups in sentences.
\cite{prabhakaran2019} takes a similar approach, but manipulates proper names in text, and uses sensitivity measures
which are similar to ours.
Our work can be understood as the image analog of theirs, with one key difference: rather than manually perturbing the input data, we utilize a generative model to manipulate face characteristics in the images. Our counterfactual face manipulation techniques are related to that of \cite{mcduff2018},
but whereas they use 3D simulations to generate images, we use Generative Adversarial Networks \cite{gans}.

Our approach is most similar to \cite{Rodden}, which explores how the output of a gender classifier changes in response to the alteration of various characteristics of the same person (e.g.,~hair length, facial expression, etc.).
Our approach differs in that we use generative techniques to systematically vary characteristics of a face image, rather than relying on a manually curated set of real images.
Also, we prefer to look at smiling classifiers, as these have more beneficial applications and
fewer harmful ones, as described in Section~\ref{secc:ethical}.

\section{Methods}
\label{sec:methods}
We now outline our \textit{counterfactual face attribute manipulation} framework, which is motivated by questions of the form: how would the prediction change if a single factor of variation in the image were altered?
\subsection{Generative model}

Generative adversarial networks (GANs) \cite{gans} are a framework for learning generative models whereby two networks are trained simultaneously in an adversarial manner. The generator $G$ takes as input a latent code vector $\z \in \mathcal{Z}$ sampled from a fixed prior $p(z)$ and outputs an image $\x \in \mathcal{X}$. The discriminator $D$ takes as input an image $\x \in \mathcal{X}$ that is sampled from either the training set or the generative distribution defined by $G$. $D$ is trained to discriminate between images from these two distributions while $G$ is trained simultaneously to generate samples that
fool $D$ into thinking they come from the data distribution.

In this work, we utilize the architecture and progressive training procedure introduced in \cite{pggan} and train with the Wasserstein GAN loss \cite{wgan} with a gradient penalty \cite{Gulrajani2017}:

\begin{equation}
\begin{aligned}
    \min_G \max_D &\mathbb{E}_{\x \sim p(\x)} [D(\x)] - \mathbb{E}_{\z \sim p(\z)} [D(G(\z))] \\
    & + \lambda \mathbb{E}_{\x \sim p_I(\x)}[(||\nabla_x D(\x) ||_2-1)^2]
\end{aligned}
\end{equation}
Here, $p(\x)$ is the data distribution, $p(\z)$ is the prior noise distribution and $p_I(\x)$ is  given by sampling uniformly along lines between points in the true and generative distributions.

The basic GAN framework lacks an inference mechanism, i.e., a method of identifying the latent code $\z$ that generated a particular image $\x$. Inference is central to our method since we  use annotated data to estimate directions in latent space that correspond to different factors of variation in the dataset. In order to obtain an approximate inference mechanism we train an encoder $E$ to map from images to latent codes. Specifically, for a fixed generative model, we train the encoder with an $\ell_2$ loss to predict the latent code $\z$ that generated $\x = G(\z)$:
\begin{align}
\min_E \mathbb{E}_{z \sim p(z)} ||z - E(G(z))||_2
\end{align}

\begin{figure}[t!]
  \centering
    \includegraphics[width=\linewidth]{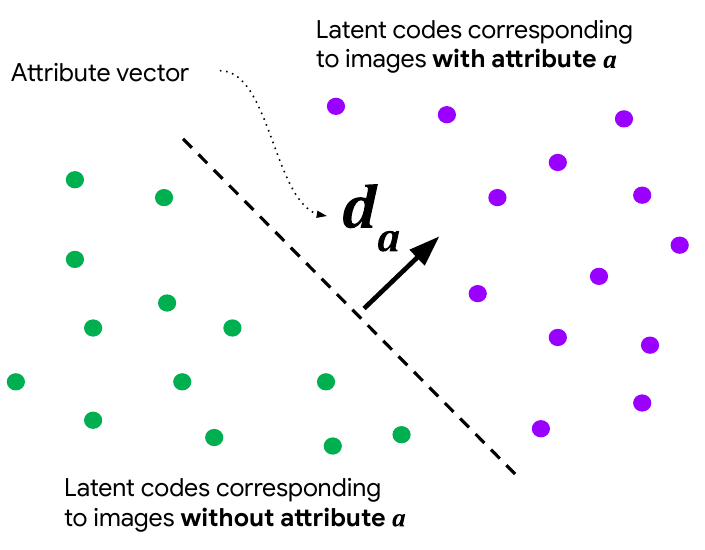}
  \caption{Attribute vectors are computed by training a linear classifier to discriminate between latent codes corresponding to images with and without a given attribute $a$. The attribute vector $\mathbf{d_a}$ is taken to be the vector orthogonal to the decision boundary, normalized to have unit $\ell_2$ norm.}
\end{figure}
\label{fig:classifier}

\subsection{Face Attribute Vectors}\label{secc:attributes}
\begin{figure}[t!]
  \centering
  \includegraphics[width=\linewidth]{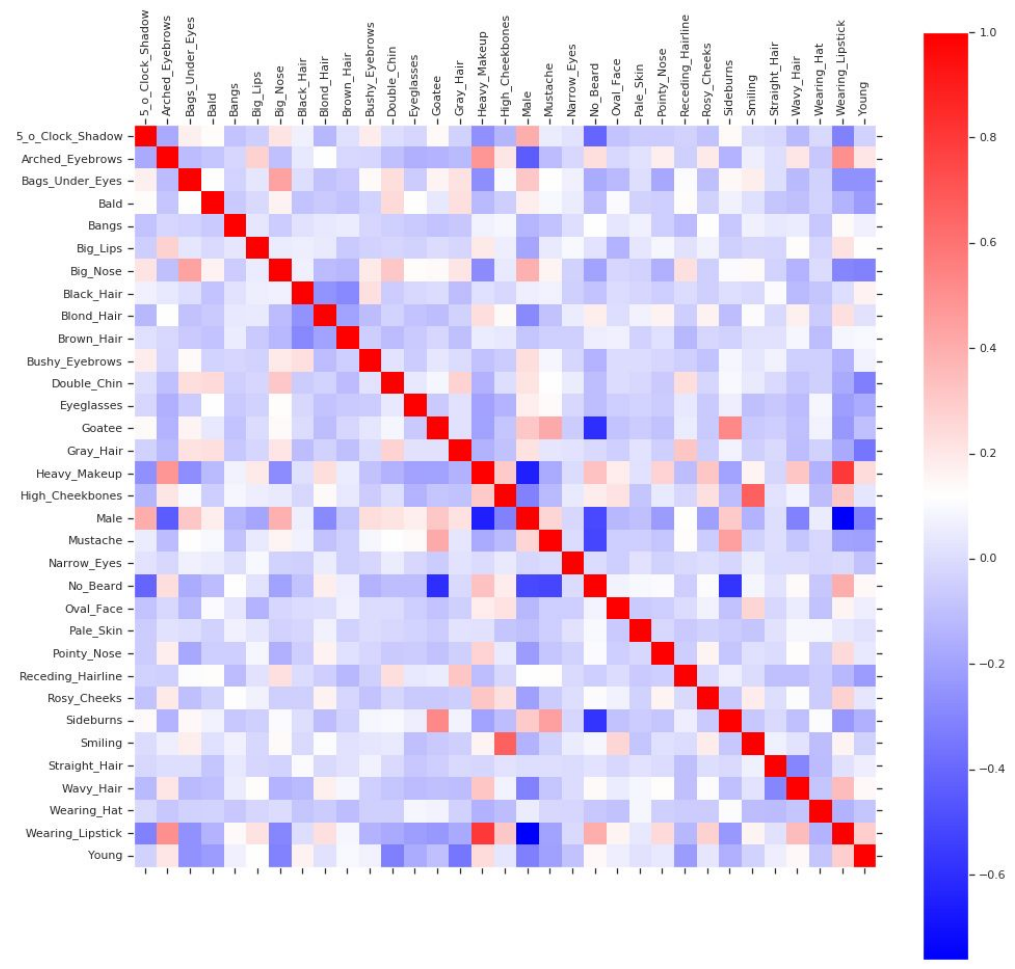}
\caption{Empirical correlation matrix of CelebA attributes.}
\label{fig:celeba_attribute_correlation}
\end{figure}

\fig{celeba_attribute_correlation} shows the correlation matrix of CelebA attributes, computed from the training images.

\begin{figure*}
  \centering
  \subfloat[]{%
       \includegraphics[width=0.45\linewidth]{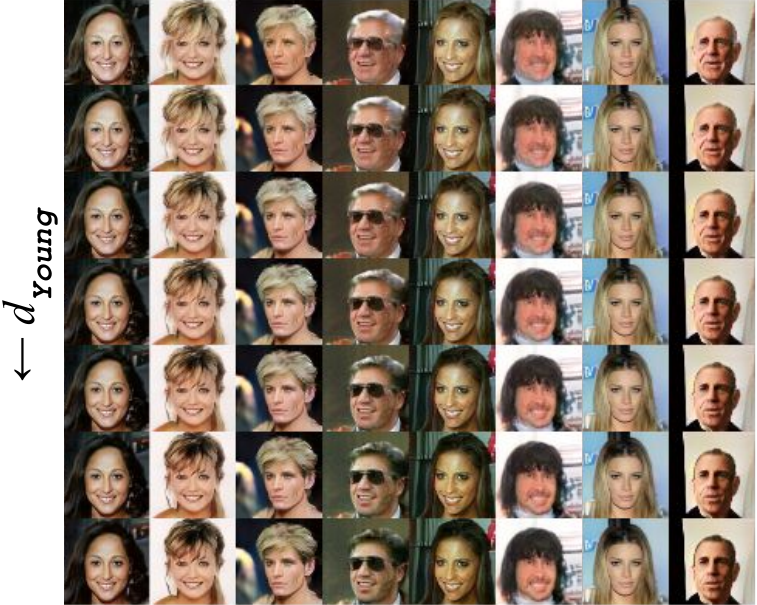}}
       \qquad
  \subfloat[]{%
       \includegraphics[width=0.48\linewidth]{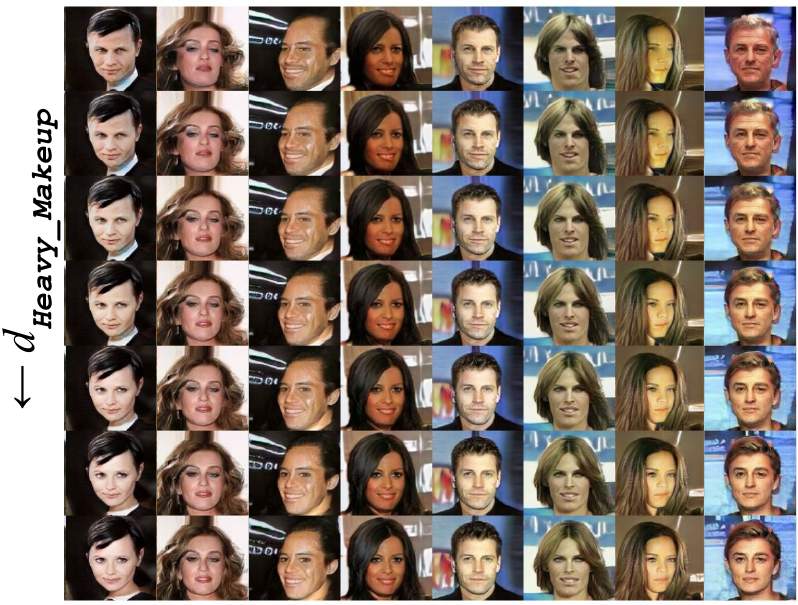}}
  \caption{Traversing in latent code space along the (a) \texttt{Young} and (b) \texttt{Heavy\_Makeup} attribute direction. The center row shows images generated from random $z \sim p(z)$. Rows moving down (up) show images generated by traversing the \texttt{Young}/\texttt{Heavy\_Makeup}  attribute vector in a positive (negative) direction. We see age/makeup related facial characteristics being smoothly altered while the overall facial expression remains constant.}
    \label{fig:makeup_young_ex}
\end{figure*}

Inspired by \cite{tcav}, we compute attribute vectors by training a linear classifier to separate the latent codes inferred from images annotated with the attribute from latent codes inferred from images annotated without the attribute. We take the attribute vector $\mathbf{d_a}$ to be the vector orthogonal to the linear classifier's decision boundary. We normalize all attribute vectors to have unit $\ell_2$ norm.


In our formulation, we examine the effect of a range of CelebA attributes (e.g. \texttt{Young}, \texttt{Heavy\_Makeup}, \texttt{No\_Beard}, etc.) on \texttt{Smiling} predictions.
Since the \texttt{Smiling} attribute is highly correlated with several other attributes in CelebA (see \fig{celeba_attribute_correlation} for the full attribute correlation matrix), we ensure the \texttt{Smiling} attribute is equally represented in both the positive and negative classes of the other attributes we model to avoid inadvertently encoding it in the attribute vector. We also balance the \texttt{Male} attribute in an attempt to disentangle the set of attributes highly correlated with it. 

A face attribute vector $\mathbf{d_a}$  can be traversed, positively or negatively, to generate images that smoothly vary along the factor of variation defined by the attribute. For example, given an image $\x = G(\z)$, we can synthesize new images that differ along the factor of variation defined by $\mathbf{d_a}$ via $G(\z + \lambda\mathbf{d_a})$. We consider $\lambda \in [-1, 1]$ and note that as $\lambda$ increases above 0, the attribute will be present in increasing amounts. Similarly, as $\lambda$ decreases below 0, the attribute will be removed. Despite being operationalized as binary within the CelebA dataset, the vast majority of characteristics captured by the annotations are continuous. Thus, this continuous traversal of attribute vectors gives rise to perceptually meaningful alterations in the images. For example, \fig{makeup_young_ex}(a) demonstrates images synthesized by perturbing $z's$ along the $\mathbf{d}_{\texttt{Young}}$ direction. We emphasize again that many of the annotations are subjective, and thus, these  manipulations reflect how the particular attributes were operationalized and annotated within the CelebA dataset.


\subsection{Score Sensitivity Analysis}\label{secc:score_sensitivity}
We now describe our method of testing the sensitivity
of the classifier to counterfactual face attribute manipulation.
Let $f$ denote a trained classifier that takes an image $x \in \mathcal{X}$ as input and outputs a continuous value
in the range $[0, 1]$.
We are interested in how $f(x)$ is affected as the input $\x$ changes in a controlled manner.
When $\x$ is generated by a process $G$, i.e.\ $\x = G(\z)$, we are similarly interested in how
$f \circ G (\z)$, i.e.\ $f(G(\z))$, is affected as $\z \in Z \subset \mathcal{Z}$ changes in a controlled manner.

With this in mind, we define the {\em score sensitivity} of $f \circ G$ to a vector $\dvec$, calculated over $Z$, as:\footnote{This is closely related to
error stability in learning theory \cite{bousquet2002}.
We believe that sensitivity analysis analogous to what \cite{bousquet2002} call ``hypothesis stability'', namely $\mathbb{E}_Z [|f(G(\z + \dvec)) - f(G(\z))|]$
can also be a useful form of model evaluation, however
it is less related to the forms of systematic model bias that we are concerned with.}
\
\begin{align}
    S(f \circ G, \dvec; Z) = \mathbb{E}_Z [f \circ G(\z + \dvec) - f \circ G(\z)]
\end{align}
\
In our experiments below, $Z$ will be sampled randomly from $\mathcal{Z}$,
using a prior $p(\z)$. However, one might alternatively obtain $Z$ from a set
of images $X \subset \mathcal{X}$  using the encoder $E$, i.e., $Z=E(X)$.
In practice, we will drop the parameter $Z$ when it is clear from context.
In other words, $S(f \circ G, \d)$ compares an image classifier's output given an image $G(\z)$ with the classifier's output given the image manipulated by an attribute vector $G(\z+\dvec)$.

For a face attribute vector $\mathbf{d_a}$,
$S(f \circ G, \mathbf{d_a})$ will be close to $0$ only if the classifier $f$ is blind to the
differences between images $G(\z + \mathbf{d_a})$ and $G(\z)$.
In contrast, if $S(f \circ G, \d)$ has a large magnitude,
there are two possible explanations.
Firstly, $f$ could be sensitive to pixel representations of attribute $a$ in image space.
In other words, if $a$ represents
something relevant to the classification task (e.g., mouth shape in the context of smiling classification) a large $S(f \circ G, \mathbf{d_a})$ is expected. However,  if $a$ is a characteristic that we believe should be irrelevant to classification (e.g., hair color in the context of smiling classification) then a large mangnitude $S(f \circ G, \mathbf{d_a})$ indicates the potential of an undesirable bias in the model.
Secondly, it may be that the images $G(z)$ and $G(z+\mathbf{d_a})$ differ in dimensions that are not relevant to the attribute $a$, and $f$ is sensitive to those dimensions of variation.

\subsection{Attribute Perception Sensitivity Analysis}\label{secc:perceptual_sensitivity}

In order to tease apart the two possibilities listed above,
it is useful to consider
whether humans interpret the differences between $G(\z)$ and $G(z+\mathbf{d_a})$ to be
relevant to the attribute $a$.
In this paper we will refer to such interpretations as ``perceptions''.
For example, when we manipulate the attribute \texttt{Young}, we find that hair colour or baldness often change,
which is appropriate (Figure~\ref{fig:makeup_young_ex}a).
Conversely, it is possible that humans might perceive differences to $G(\z)$ and $G(\z+\mathbf{d_a})$ that relate to
dimensions of facial variation other than $a$.
For example, we have found that in practice the hair style and/or color often changes when manipulating
the \texttt{Heavy\_Makeup} attribute (Figure~\ref{fig:makeup_young_ex}b).
We will use
    $Perceive_{a'}(G(\z), G(z+\mathbf{d_a})) \in \mathbb{R}$ to represent the
    perceptual difference in attribute $a'$ when comparing the images $G(z)$
    and $G(\z+\mathbf{d_a})$.
Analogous to score sensitivity, we define the \textit{$a'$-perception sensitivity} of $Perceive_{a'} \circ G$ to an attribute vector $\mathbf{d_a}$ as
\
\begin{align}
    S(Perceive_{a'} \circ G, \mathbf{d_a}; Z) = \mathbb{E}_Z [Perceive_{a'}(G(\z), G(\z+\mathbf{d_a}))]
\end{align}
\
One option when eliciting perceptual judgments from human subjects, is to simply ask
just ask whether they perceive any difference in $a'$ between the images $G(\z)$ and $G(\z+\mathbf{d_a})$, i.e.,
$Perceive_{a'}(G(\z), G(\z+\mathbf{d_a})) \in {0, 1}$.

\subsection{Classification Sensitivity Analysis}\label{secc:classification_sensitivity}

Given a threshold, $0 \leq c \leq 1$, binary classifications can be obtained from the classifier $f$:
\begin{align}
y(\x) = \mathbb{I}[f(\x)  \geq c] \in \{0, 1\}
\end{align}
\
Directly analogous to the score sensitivity above for the function $f$, we can define the {\em classification sensitivity} of the model towards $\dvec$,
calculated over $Z$, as:
\
\begin{align}
    S(y \circ G, \dvec; Z) = \mathbb{E}_Z [y \circ G(\z + \dvec) - y \circ G(\z)]
\end{align}
\
Again, when $Z$ is clear from context we will omit it.
That is, $S(y \circ G , \dvec)$ measures the likelihood of a classification flipping,
either from `not smiling' to `smiling' or vice versa, when $\z$ is perturbed by $\dvec$.

When auditing a smiling classifier for bias, it is useful to be able to drill down further and consider the
changes from `not smiling' to `smiling' distinct from changes from `smiling' to `not smiling'.
To do this, we restrict the subset $Z \in \mathcal{Z}$ over which the expectations are calculated
to either $Z_0=\{\z | y(G(\z))=0\}$ or $Z_1=\{\z | y(G(\z))=1\}$, respectively.
As shorthands for the restrictions of classification sensitivity, we will use the notation $S^{0 \rightarrow 1}(y \circ G | \dvec)$ and
$S^{1 \rightarrow 0}(y \circ G | \dvec)$, respectively.

\section{Results}
\label{Results}
CelebA \cite{celeba} is a dataset of celebrity faces annotated with the presence/absence of 40 face attributes. We train a convolutional neural network to predict the \texttt{Smiling} attribute on 128x128 images. We train  with a cross-entropy  objective and use a threshold of $c = 0.5$ for final predictions. Error statistics, disaggregated by a select set of attributes, are given in \tab{celeba}.

\begin{table}
   \small 
   \centering 
   \begin{tabular}{l|ccc} 
    Data split& \textbf{Accuracy} & \textbf{FPR} & \textbf{FNR} \\
       \hline
   \hline
   Total & 89.330\% & 14.416\% & 6.929\% \\
   \hline
   \texttt{Young=0} & 87.170\% & 17.727\% & 8.193\% \\
   \texttt{Young=1} & 90.022\% & 13.391\% & 6.509\% \\
   \hline
   \texttt{Male=0} & 89.962\% & 16.548\% & 4.984\% \\
   \texttt{Male=1} & 88.356\% & 11.948\% & 11.190\%\\
   \hline
   \texttt{Heavy\_Makeup=0} & 88.735\% & 12.346\% & 9.769\% \\
   \texttt{Heavy\_Makeup=1} & 90.203\% & 19.046\% & 4.098\% \\
   \hline
   \texttt{Wearing\_Lipstick=0} & 88.254\% & 12.071\% & 11.267\% \\
   \texttt{Wearing\_Lipstick=1} & 90.315\% & 17.521\% & 4.195\% \\
   \hline
   \texttt{No\_Beard=0} & 88.942\% & 10.683\% & 11.699\% \\
   \texttt{No\_Beard=1} & 89.396\% & 15.263\% & 6.352\% \\
    \hline
   \texttt{5\_o\_Clock\_Shadow=0} & 89.303\% & 14.739\% & 6.823\% \\
   \texttt{5\_o\_Clock\_Shadow=1} & 89.569\% & 12.014\% & 8.128\% \\
   \hline
   \texttt{Goatee=0} & 89.363\% & 14.667\% & 6.731\% \\
   \texttt{Goatee=1} &  88.634\% & 10.500\% & 13.016\% \\
    \hline
   \end{tabular}
   \caption{CelebA smiling attribute classification results.}
   \label{tab:celeba}
\end{table}

We train a progressive GAN on 128x128 CelebA images with a 128-dimensional zero-mean Gaussian prior. For each attribute in the CelebA dataset, we estimate the directions in latent code space corresponding to these factors of variation  via the procedure described in \secc{attributes}. We infer 800 latent codes for each of the positive (attribute is present) and negative (attribute is not present) classes. We select a random 80\% of the vectors for training and evaluate classification accuracy on the remaining 20\%.
The classification accuracy of each linear classifier is shown in \fig{class_acc} in the Appendix.

\begin{figure*}[t!]
  \centering
  \includegraphics[width=\textwidth]{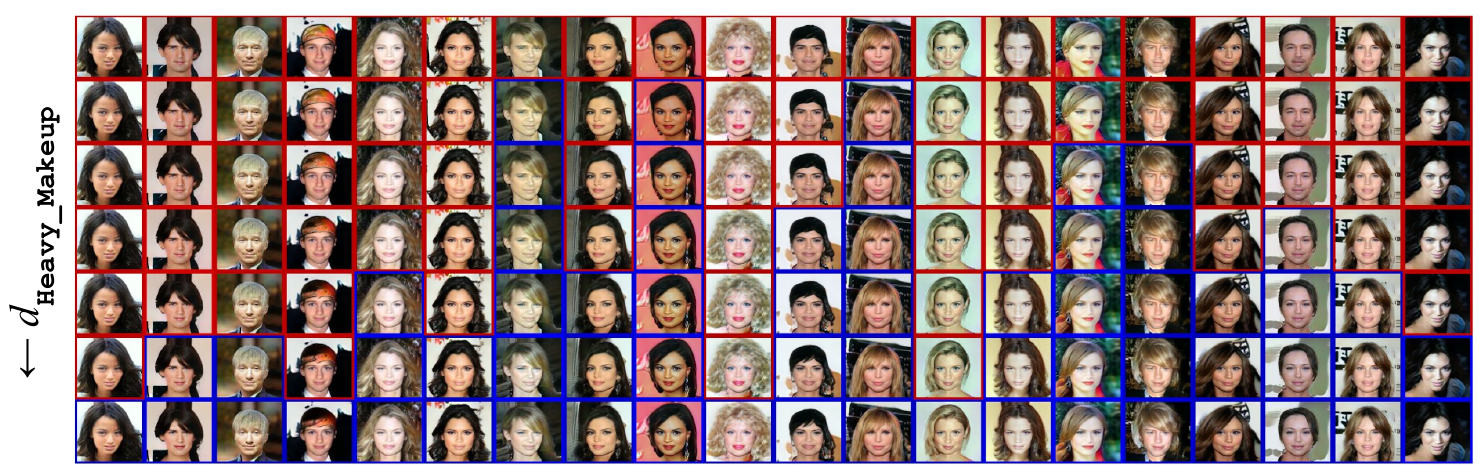}
  \includegraphics[width=\textwidth]{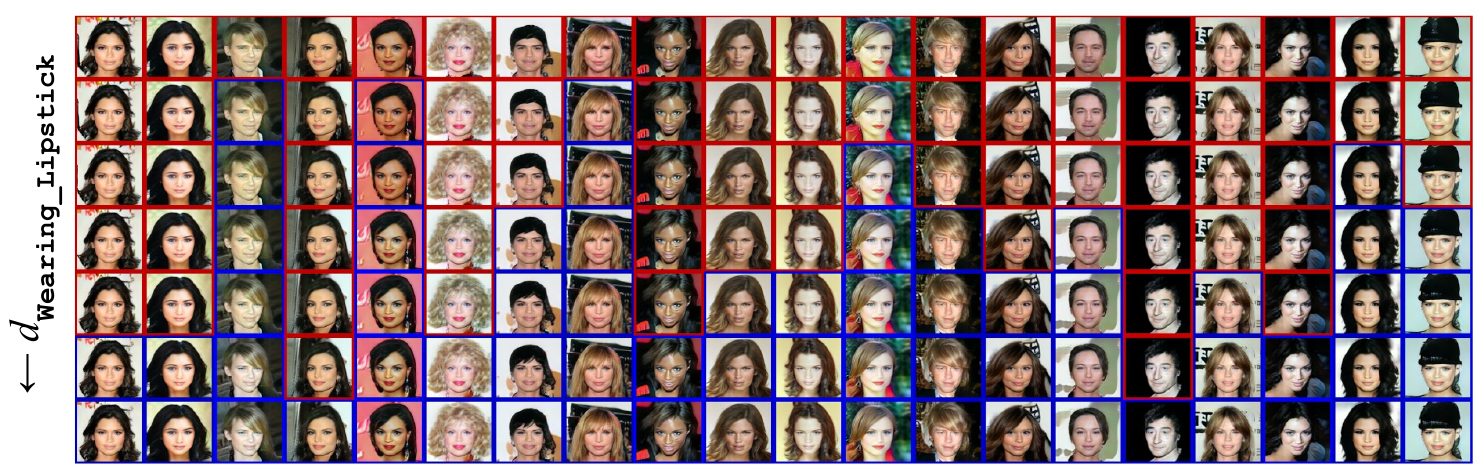}
  \includegraphics[width=\textwidth]{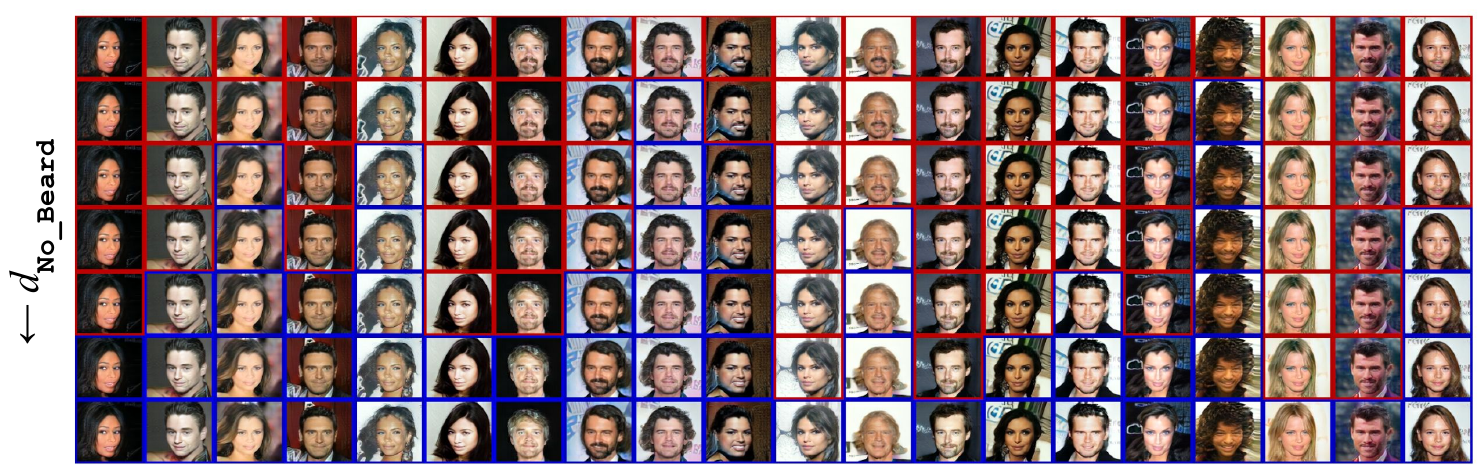}
\caption{Examples where smiling classifier's prediction flips when traversing different attribute vectors. Blue (red) boxes indicate a smiling (not smiling) prediction. }
\label{fig:img_flips}
\end{figure*}

To illustrate the visual effect of the generative face attribute manipulation, \fig{makeup_young_ex} shows
samples generated from the model by traversing the latent code space along the $\mathbf{d}_{\texttt{Heavy\_Makeup}}$ and $\mathbf{d}_{\texttt{Young}}$ directions respectively.
We see age/makeup related facial characteristics being smoothly altered while the overall facial expression remains constant. Further examples, can be found in \fig{img_flips}, and the examples in both Figures were
created by sampling randomly from the prior $p(z)$ on $\mathcal{Z}$. We see that the faces and
the manipulations are generally realistic, a topic we'll discuss more in Section~\ref{sec:social_validity}.

Many of the CelebA attributes are highly correlated in the dataset (see \fig{celeba_attribute_correlation}  for the full correlation matrix). The GAN learns to separate some factors (as evidenced, for example, by the consistent facial expressions in \fig{makeup_young_ex}), but does not cleanly disentangle all factors of variation in human faces. For example, there are many facial attributes that, through the societal production of the gender binary, have come to signify traditionally masculine or feminine characteristics. Social norms  around gender expression lead these features to be highly correlated in society and the CelebA dataset reflects, if not amplifies, these correlations. Empirically, we observe these attributes remain entangled to differing degrees in the generative model. Consequently, traversing one attribute vector (e.g.~\texttt{Heavy\_Makeup}) sometimes changes a larger set of facial characteristics (e.g.~removing/adding a beard, altering hair length, etc.) that tend to be correlated with the attribute in question. Future work is aimed at further disentangling factors of variation that are highly correlated in the dataset.

\begin{figure}[t!]
  \centering
  \includegraphics[width=\linewidth]{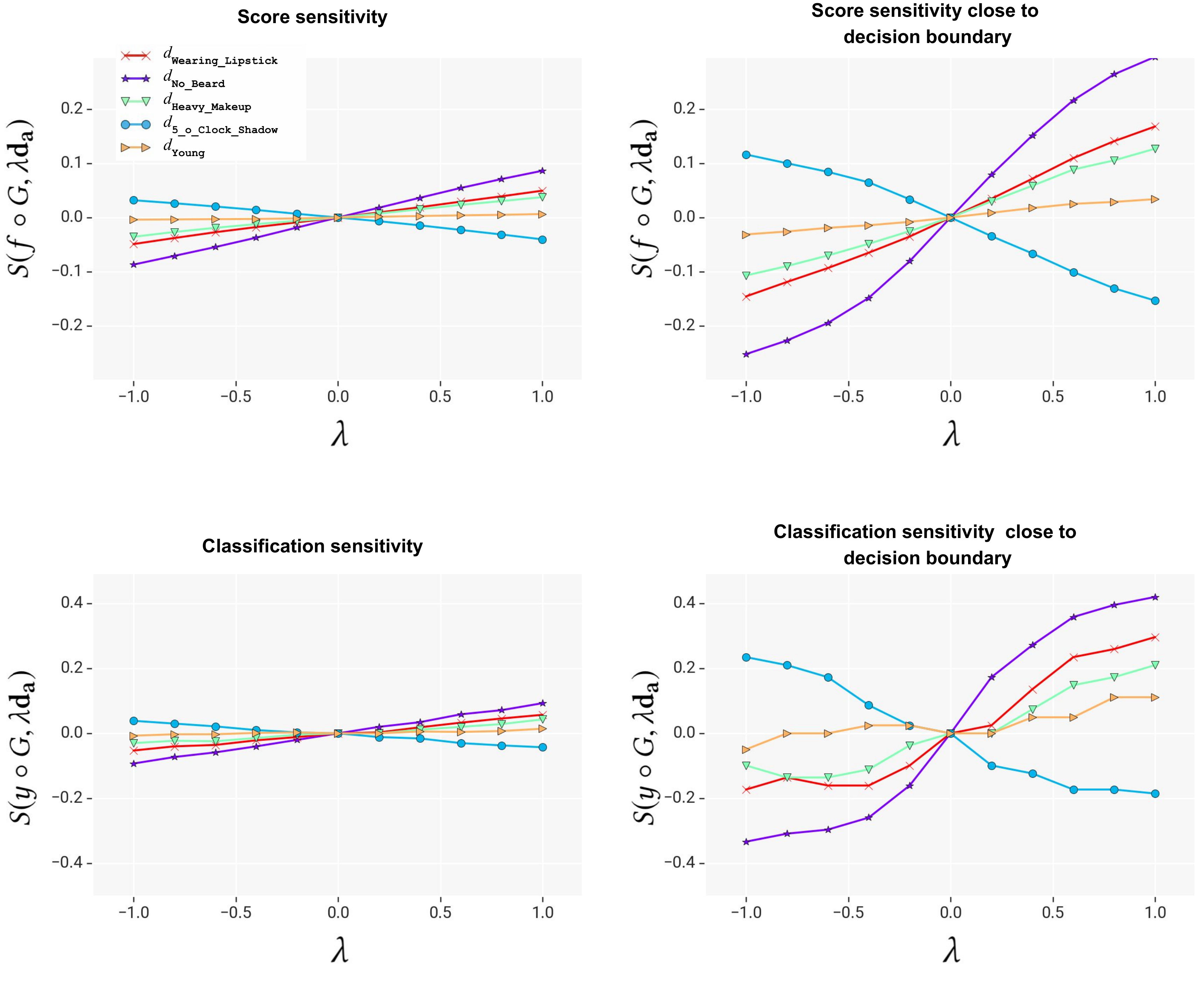}
\caption{Sensitivity analyses of the smiling classifier. For each attribute, the y-axis denotes $\lambda$ equally space within $[-1, 1]$, i.e. the degree to which the attribute has been added ($\lambda > 0$) or removed ($\lambda < 0$) in the image.
}
\label{fig:psmile_change}
\end{figure}


\begin{figure*}[t!]%
    \includegraphics[width=\linewidth]{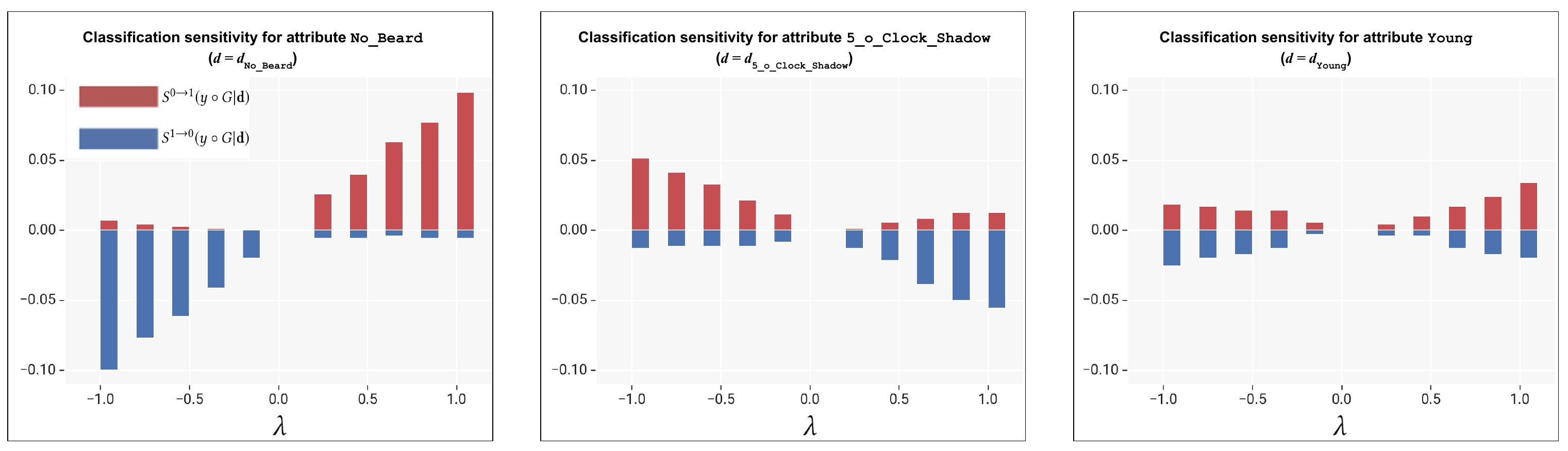}%

    \begin{minipage}[c]{0.67\textwidth}
    \includegraphics[width=\linewidth]{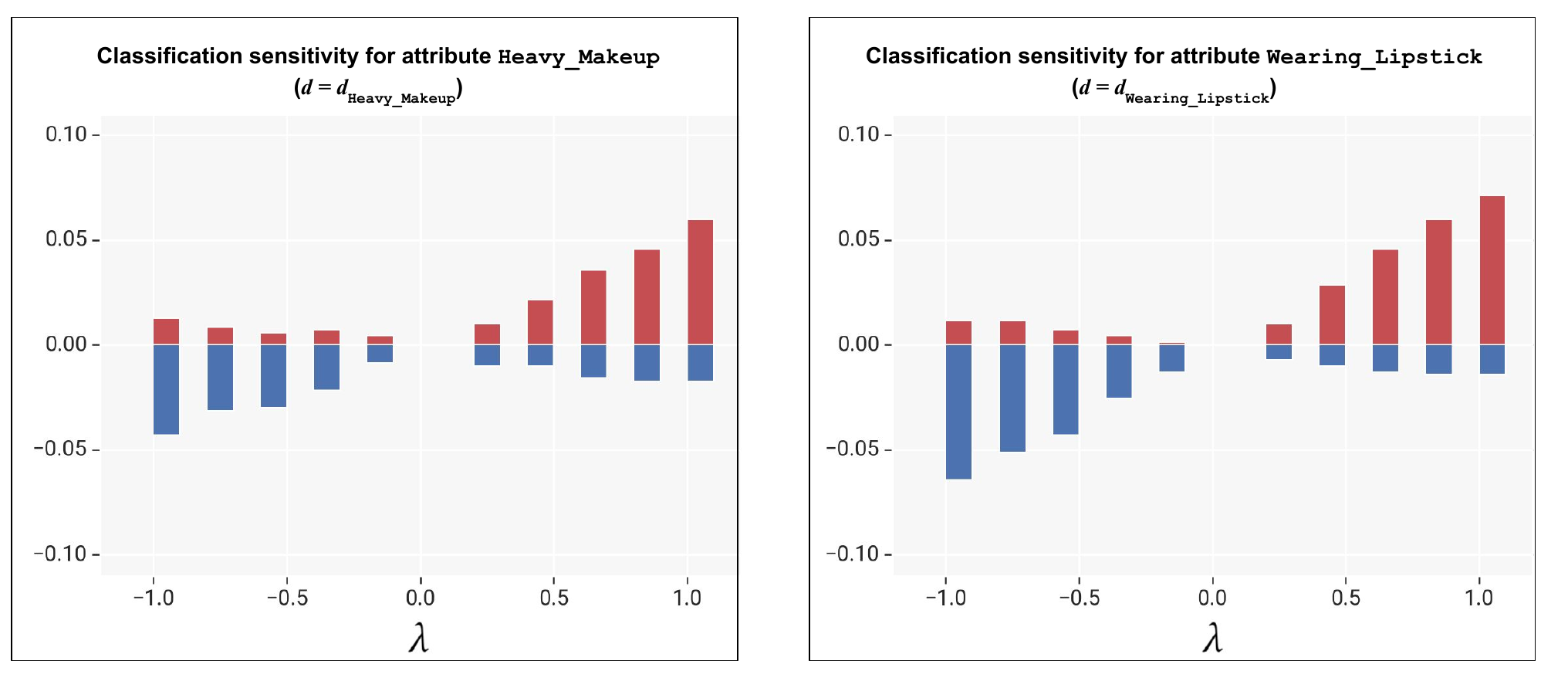}%
    \end{minipage}
    \begin{minipage}[c]{0.32\textwidth}
      \caption{Effect of traversing different attribute vectors on the smiling classifier's binary prediction. $S^{0\rightarrow1}$ indicates frequency with which the binary smiling classification flips from 0 to 1 and  $S^{1\rightarrow0}$ indicates frequency with which the binary smiling classification flips from 1 to 0.}%
    \end{minipage}

\end{figure*}
\label{fig:MFL}%

Our approach involves traversing directions in latent code space to generate images that vary smoothly along some factor of variation. We performed a simple sanity check to ensure that images generated from nearby latent codes are consistently classified by the smiling classifier. First, we construct a large set of generated images that are classified as smiling. Then we generate new images by sampling latent codes uniformly from lines connecting two different codes of the generated images. We find that 98\% of the new images are also classified as smiling. We similarly  test this for images classified as not smiling and find the same results.

Next, we assess the score sensitivity of the smiling classifier to the counterfactual face attribute manipulation, using the measure proposed in Section~\ref{secc:score_sensitivity}.
 Specifically, we use
different multiples of attribute vector $\mathbf{d_a}$ to manipulate attribute $a$ in an image. For example, given image $\x = G(\z)$, we can increase the degree of makeup on the face by generating
$$G(\z + \lambda \mathbf{d}_{\texttt{Heavy\_Makeup}}), \lambda \in (0, 1].$$
Conversely, generating
$$G(\z + \lambda \mathbf{d}_{\texttt{Heavy\_Makeup}}), \lambda \in [-1, 0)$$ will decrease the amount of makeup on the face in the image.

\fig{psmile_change} (top-left) plots the score sensitivity $S(f \circ G, \lambda \mathbf{d_a} )$ for five attributes $a$ in the CelebA dataset and for $\lambda \in [-1, 1]$. Interestingly, we observe that linear changes to $\z$ result in
roughly linear changes in the scores output by the classifier.
\fig{psmile_change} (top-right) restricts the domain of the sensitivity analysis to $\z$ for which $G(z)$
is near the decision boundary of the smiling classifier. Specifically, $0.3<f(G(\z))<0.7$.
We see both greater sensitivity generally (compared to top left), as well as greater differences in the
sensitivities to individual attributes.

\fig{psmile_change} (bottom-left, bottom-right) show the corresponding analyses using classification sensitivity.
As expected, the classification sensitivity is much larger than the score sensitivity when $G(z)$ is
near the decision boundary (note the difference in scale of $y$-axes between top-right and bottom-right).
While \fig{psmile_change} (bottom-left, bottom-right) captures the systematic directional sensitivity of
the model to the attributes $\mathbf{d_a}$, it doesn't capture the magnitude of the number of flips from
`smiling' to `not smiling' and vice versa. These are shown in Figure \ref{fig:MFL}.

A fair smiling classifier should also perform consistently regardless of the gender or gender expression of the individual in the photo.
The CelebA \texttt{Male} attribute captures something related to these concepts. However, as discussed in \secc{ethical}, we chose not to use the \texttt{Male} annotations. We emphasize that image manipulations based on the  \texttt{Male} attribute would not alter anything inherent to gender identity or gender expression. Rather, these manipulations would simply reflect a specific set of correlated features that occur in different proportions in the images annotated $\texttt{Male}=0$ and $\texttt{Male}=1$.

\fig{psmile_change} suggests the smiling classifier is sensitive to different attributes. We now turn to visual inspection to (i) assess the perceptual effect of traversing the attribute vectors and (ii) ensure the basic mouth/facial expression that signifies the presence/absence of a smile does not change when traversing the attribute vectors. \fig{img_flips} shows several examples of images generated by traversing different attribute vectors. Blue or red boxes outline each image to indicate the smiling classifier's prediction for the image. Red boxes indicate a not smiling prediction and green boxes indicate a smiling prediction.  We observe the basic facial expression remains constant when traversing many of the attribute vectors (e.g., \texttt{Young}, \texttt{Blond\_Hair}, \texttt{5\_o\_Clock\_Shadow}, \texttt{Heavy\_Makeup}), despite other characteristics of the face changing. This indicates the generative model has sufficiently disentangled the factor of variation defined by the CelebA attribute from the features of a smile. However, despite the faces (perceptually) having nearly identical mouth expressions, the classifier is nonetheless sensitive to the changes.


\section{Discussion: Social Validity of Counterfactual Images}
\label{sec:social_validity}
The results discussed in Section~\ref{Results} illustrate some fundamental challenges in
counterfactual audit studies involving images. In discussing these challenges, it is useful
to be clear about the different ways in which attributes are represented in our counterfactual study:
\\
\textbf{Data}: Attributes are explicitly represented as binary annotations on images, and pairs of attributes
are correlated to various degrees (see \fig{celeba_attribute_correlation}).
\\
\textbf{Latent space}: Attributes are represented as vectors in the latent space, which is the input to the GAN. Pairs of attributes have different angles between them.
\\
\textbf{Human perception}: When humans interpret images, they can report perceiving
attributes, and can also report perceiving changes to attributes.  For example, this includes perception of color and shape.
\\
\textbf{Constructs}: The cognitive representations of attributes are entwined in a web of associative relationships, both with other attributes and other concepts, including causal, semantic and affective relationships (e.g., \cite{greenwald1998measuring, oh2019revealing}). For example, this includes perceptions of age and gender presentation
(e.g., \cite{olivola2010elected, le2002evaluating}).
\\
\textbf{ML Models}: Classifiers may implicitly learn certain attributes from their training data.
Interpretability techniques such as Testing with Concept Activation Vectors can reveal
a model's implicit dependence on an attribute \cite{tcav}.
\\
\indent Critical to the program of testing models using counterfactual images is
a sophisticated understanding of how these
different representations of attributes interact with each other. We note that causal relationships between constructs form a critical part
of our cognitive representations of face attributes. Importantly, the latent space in which we perform
attribute perturbations contains no model of causation, and we caution against attempts
to understand our experiments in terms of causal relationships between attributes.
It is as nonsensical to say that our process has learned that aging
often causes changes to hair color as it would be to say that the model has learned that changes to hair color
often cause aging. Similarly, make up does not cause changes to hair; the latent space has simply learned
various relationships between hair styles and the presence or degree of makeup.

In our case, using a GAN
to generate our images,  one aspect of experimental validity that arises is that $G(z)$ should be
a realistic image. We describe this as \textit{social validity} in order to acknowledge that images are
inherently sociotechnical artifacts.
Additionally, we require both that $G(\z + \mathbf{d_a})$ have social validity, and
that the difference between images $G(\z)$ and $G(\z + \mathbf{d_a})$ corresponds ``validly''
to differences in the human perception of $a$ in the images.

In the definition of counterfactual fairness presented by \cite{Kilbertus2017, Kusner2017},
the fairness of a model is determined by assessing the effect of counterfactual interventions on a sensitive
variable (e.g., race or gender) over a causal graph. In our case, the sensitive variables are
human interpretations of attributes in images, which we manipulate only indirectly, via
the attribute vectors $\mathbf{d_a}$ in latent space. For this
counterfactual analysis to be valid,  we require that
manipulating $\mathbf{d_a}$ have no causal impact on the perception of smiling.
Our own visual analysis of randomly generated images supported this, although testing this more rigorously at
large scale and with a large pool of human subjects remains future work.
However, we have found that were often perceptible differences
in hair style and/or color when manipulating
various non-hair attributes in the latent space (e.g., the manipulation of
the \texttt{Heavy\_Makeup} attribute creates some perceptible differences in hair
in Figure~\ref{fig:makeup_young_ex}b).
This illustrates the point that in practice attributes are frequently non-orthogonal in the latent space,
no doubt due to their correlations in the data (\fig{celeba_attribute_correlation}).

These points all suggest the need for a research agenda which explores two things.
Firstly, there is a need to understand the relationships
between the latent spaces over which manipulations are performed, on the one hand, and human perception
of socially relevant attributes, on the other. To do this, we require research studies involving human participants and how they perceive differences in images that result from manipulating the latent space.
Such experiments would enable the sketch of an attribute perception sensitivity function given in
Section~\ref{secc:perceptual_sensitivity} to be utilized, enhancing the
conclusions that may be drawn from the sensitivity analyses.
Secondly, better understanding of our cognitive models of face attributes---and in particular the causal relationships between them---is required to understand the validity of the ``interaction'' effects that we
see whereby manipulating one attribute vector alters the perception of other attributes.

\section{Conclusion}

We have proposed a technique to discover unintended biases present in facial analysis models by leveraging generative adversarial networks. The approach provides fine-grained, controlled manipulation of face images to reveal hidden and unintended biases in how faces are conceptualized by the model, and how this affects classifier decisions. To accompany this approach, we provide metrics that aid in evaluating the sensitivity of a facial analysis classifier to counterfactual face attribute manipulation---in terms of how the score of the classifier changes as well as how the classification itself changes.

This proof-of-concept demonstrates potential ways generative models could survey bias, fairness and other ethical considerations, opening up the opportunity to explore the applicability of generative models across other image classification problems as well as on a larger scale.

We now highlight several future research directions:
\begin{itemize}
    \item Leverage human raters to further understand potential bias embedded in the attribute vectors. This is important to help distinguish the classifier bias  from potential bias in the generator / attribute annotations.
    \item Leveraging the same generative techniques to visualize dataset bias.
    \item Training the generative model on a larger and more diverse set of faces than is available for the classifier.
    \item Exploring the interaction effects of attributes on the model \cite{saltelli2010,czitrom1999}.
\end{itemize}

{\small
\bibliographystyle{ieee_fullname}
\bibliography{main}
}

\appendix
\section{Appendix}\label{sec:appendix}

\begin{figure}[h!]
  \centering
  \includegraphics[width=0.9\linewidth]{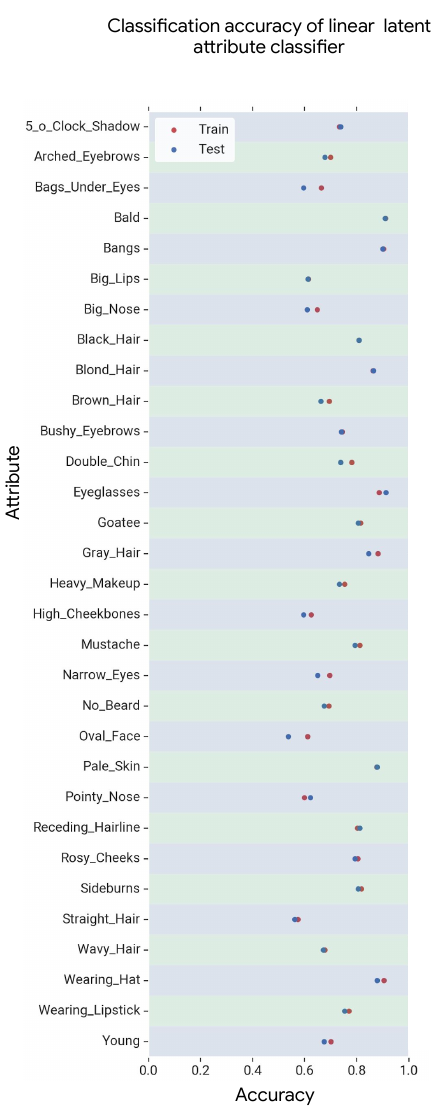}
\caption{Classification accuracy of each linear classifiers trained to separate latent codes computed from images annotated with and without the attribute. Recall, the vector orthogonal to the decision boundary gives the attribute vector.}
\label{fig:class_acc}
\end{figure}

\end{document}